%% file: main.tex
\documentclass{article} % For LaTeX2e
\usepackage{iclr2018_conference,times}
\usepackage{hyperref}
\usepackage{url}
\usepackage{booktabs}       % professional-quality tables
\usepackage{amsfonts}       % blackboard math symbols
\usepackage{nicefrac}       % compact symbols for 1/2, etc.
\usepackage{microtype}      % microtypography
\usepackage{amsmath}
\usepackage{amsthm}
\usepackage{color,mathtools}
\usepackage{enumitem}
\usepackage{multirow}
\usepackage{graphicx}
\usepackage{subcaption}
\usepackage{bm}
\usepackage{rotating}
\usepackage{pgffor}
\usepackage{array}
\usepackage{tabularx}
\usepackage{tabu}

\graphicspath{ {images/} }

\title{Variational Inference of Disentangled Latent Concepts from Unlabeled Observations}
\author{Abhishek Kumar, Prasanna Sattigeri, Avinash Balakrishnan \\
IBM Research AI\\
Yorktown Heights, NY\\
\texttt{\{abhishk,psattig,avinash.bala\}@us.ibm.com} \\
}

\newcommand{\bvae}{$\beta$-VAE }
\newcommand{\bvaenosp}{$\beta$-VAE}
\newcommand*\rot{\rotatebox{90}}
\newlength\mytemplength

\iclrfinalcopy % Uncomment for camera-ready version, but NOT for submission.
\input{notation}

\begin{document}
\maketitle

\input{abstract}

\input{intro}

\input{formulation}

\input{experiments}

\input{related}

\input{conclusion}

%\subsubsection*{Acknowledgments}

\bibliography{ml}
\bibliographystyle{iclr2018_conference}
\newpage

\input{appendix}

\end{document}

%% file: notation.tex
\newcommand{\punt}[1]{}

\def\argmax{\mathop{\rm arg\,max}}
\def\argmin{\mathop{\rm arg\,min}}

\newcommand{\expect}{\mathbb{E}}

\newcommand{\kl}{\textrm{KL}}

\def\argmax{\mathop{\rm arg\,max}}
\def\argmin{\mathop{\rm arg\,min}}

\newcommand{\vmu}{\bm{\mu}}

\newcommand{\vI}{\mathbf{I}}

\newcommand{\vSigma}{\bm{\Sigma}}

\newcommand{\vw}{\mathbf{w}}
\newcommand{\vx}{\mathbf{x}}

\newcommand{\vz}{\mathbf{z}}

\newcommand{\vy}{\mathbf{y}}

\newcommand{\bq}{\begin{equation}}
\newcommand{\eq}{\end{equation}}
\newcommand{\ba}{\begin{eqnarray}}
\newcommand{\ea}{\end{eqnarray}}

\newcommand{\mbf}[1]{\mathbf{#1}}

\newcommand{\mcal}[1]{\mathcal{#1}}
\newcommand{\remove}[1]{}

%% file: abstract.tex
\begin{abstract}
Disentangled representations, where the higher level data generative factors are reflected in disjoint latent dimensions, offer several benefits such as ease of deriving invariant representations, transferability to other tasks, interpretability, etc. We consider the problem of unsupervised learning of disentangled representations from large pool of unlabeled observations, and propose a variational inference based approach to infer disentangled latent factors. We introduce a regularizer on the expectation of the approximate posterior over observed data that encourages the disentanglement. We also propose a new disentanglement metric which is better aligned with the qualitative disentanglement observed in the decoder's output. We %evaluate the  proposed approach using several quantitative metrics and
empirically observe significant improvement over existing methods in terms of both disentanglement and data likelihood (reconstruction quality). 
\end{abstract}

%% file: intro.tex
\section{Introduction}
\label{sec:intro}
%significance/importance of disentangled representations; followed by we propose ---
Feature representations of the observed raw data play a crucial role in the success of machine learning algorithms. Effective representations should be able to capture the underlying (abstract or high-level) latent generative factors that are relevant for the end task while ignoring the inconsequential or nuisance factors. \emph{Disentangled} feature representations have the property that the generative factors are revealed in disjoint subsets of the feature dimensions, such that a change in a single generative factor causes a highly sparse change in the representation. Disentangled representations offer several advantages -- \emph{(i) Invariance:} it is easier to derive representations that are invariant to nuisance factors by simply marginalizing over the corresponding dimensions, \emph{(ii) Transferability:} they are arguably more suitable for transfer learning as most of the key underlying generative factors appear segregated along feature dimensions, \emph{(iii) Interpretability:} a human expert may be able to assign meanings to the dimensions, \emph{(iv) Conditioning and intervention:} they allow for interpretable conditioning and/or intervention over a subset of the latents and observe the effects on other nodes in the graph.
Indeed, the importance of learning disentangled representations has been argued in several recent works \citep{bengio2013representation,lake2016building,ridgeway2016survey}. 

%recent work in the area: supervised disentanglement, unsupervised approaches: infogan, beta-vae
Recognizing the significance of disentangled representations, several attempts have been made in this direction in the past \citep{ridgeway2016survey}. Much of the earlier work assumes some sort of supervision in terms of: (i) partial or full access to the generative factors per instance \citep{reed2014learning,yang2015weakly,kulkarni2015deep,karaletsos2015bayesian}, (ii) knowledge about the nature of generative factors (e.g, translation, rotation, etc.) \citep{hinton2011transforming,cohen2014learning}, (iii) knowledge about the changes in the generative factors across observations (e.g., sparse changes in consecutive frames of a Video) \citep{goroshin2015learning,whitney2016understanding,fraccaro2017disentangled,denton2017unsupervised,hsu2017unsupervised}, 
(iv) knowledge of a complementary signal to infer representations that are conditionally independent of it\footnote{The representation itself can still be entangled in rest of the generative factors.} \citep{cheung2014discovering,mathieu2016disentangling,siddharth2017learning}.  %Hadad2017? 
%\blue{(v) number of generative factors? }. 
However, in most real scenarios, we only have access to raw observations without any supervision about the generative factors. It is a challenging problem and many of the earlier attempts have not been able to scale well for realistic settings \citep{schmidhuber1992learning,desjardins2012disentangling,cohen2015transformation} (see also, \citet{higgins2016beta}). 

Recently, \citet{chen2016infogan} proposed an approach to learn a generative model with disentangled factors based on Generative Adversarial Networks (GAN) \citep{goodfellow2014generative}, however implicit generative models like GANs lack an effective inference mechanism\footnote{There have been a few recent attempts in this direction for visual data \citep{dumoulin2016adversarially,donahue2016adversarial,ournipspaper} but often the reconstructed samples are semantically quite far from the input samples, sometimes even changing in the object classes.% as  observed in \citep{ournipspaper}
}, which hinders its applicability to the problem of learning disentangled representations. 
More recently, \citet{higgins2016beta} proposed an approach based on Variational AutoEncoder (VAE) \cite{kingma2013auto} for inferring disentangled factors. The inferred latents using their method (termed as \bvae) are empirically shown to have better disentangling properties, however the method deviates from the basic principles of variational inference, creating increased tension between observed data likelihood and disentanglement. This in turn leads to poor quality of generated samples as observed in \citep{higgins2016beta}. 

%our approach; main contributions; metrics used to evaluate
In this work, we propose a principled approach for inference of disentangled latent factors based on the popular and scalable framework of amortized variational inference \citep{kingma2013auto,stuhlmuller2013learning,gershman2014amortized,rezende2014stochastic} powered by stochastic optimization \citep{hoffman2013stochastic,kingma2013auto,rezende2014stochastic}. Disentanglement is encouraged by introducing a regularizer over the induced \emph{inferred prior}. Unlike \bvae \citep{higgins2016beta}, our approach does not introduce any extra conflict between disentanglement of the latents and the observed data likelihood, which is reflected in the overall quality of the generated samples that matches the VAE and is much better than \bvaenosp. This does \emph{not} come at the cost of higher entanglement and our approach also outperforms \bvae in disentangling the latents as measured by various quantitative metrics. We also propose a new disentanglement metric, called \emph{Separated Attribute Predictability} or SAP, which is better aligned with the qualitative disentanglement observed in the decoder's output compared to the existing metrics. 
%\blue{should we also claim better estimate of the posterior? -- try the synthetic experiment}

%% file: formulation.tex
\section{Formulation}
\label{sec:form}

%start with variational inference -> amortization -> stochastic -> VAE
We start with a generative model of the observed data that first samples a latent variable $\vz\sim p(\vz)$, and an observation is generated by sampling from $p_\theta(\vx|\vz)$. The joint density of latents and observations is denoted as $p_\theta(\vx,\vz)=p(\vz)p_\theta(\vx|\vz)$. The problem of inference is to compute the posterior of the latents conditioned on the observations, i.e., $p_\theta(\vz|\vx)=\frac{p_\theta(\vx,\vz)}{\int p_\theta(\vx,\vz)d\vz}$. %where $p_\theta(\vx)=\int p_\theta(\vx,\vz)d\vz$. 
We assume that we are given a finite set of samples (observations) from the true data distribution $p(\vx)$. In most practical scenarios involving high dimensional and complex data, this computation is intractable and calls for approximate inference. Variational inference takes an optimization based approach to this, positing a family $\mcal{D}$ of approximate densities over the latents and reducing the approximate inference problem to finding a member density that minimizes the Kullback-Leibler divergence to the true posterior, i.e., $q_\vx^* = \min_{q\in\mcal{D}}\kl(q(\vz)\lVert p_\theta(\vz|\vx))$ \citep{blei2017variational}. The idea of amortized inference \citep{kingma2013auto,stuhlmuller2013learning,gershman2014amortized,rezende2014stochastic} is to explicitly share information across inferences made for each observation. One successful way of achieving this for variational inference is to have a so-called \emph{recognition model}, parameterized by $\phi$, that encodes an inverse map from the observations to the approximate posteriors (also referred as variational autoencoder or VAE) \citep{kingma2013auto,rezende2014stochastic}. The recognition model parameters are learned by optimizing the problem $\min_{\phi} \expect_\vx \kl(q_\phi(\vz|\vx)\lVert p_\theta(\vz|\vx))$, where the outer expectation is over the true data distribution $p(\vx)$ which we have samples from. This can be shown as  equivalent to maximizing what is termed as evidence lower bound (ELBO):
\begin{align}
\hspace{-4mm}\argmin_{\phi} \expect_\vx \kl(q_\phi(\vz|\vx)\lVert p_\theta(\vz|\vx)) = \argmax_{\phi} \expect_\vx \left[\expect_{\vz\sim q_\phi(\vz|\vx)}\left[ \log p_\theta(\vx|\vz)\right]  - \kl (q_\phi(\vz|\vx)\Vert p(\vz))\right]
\label{eq:elbo}
\end{align}
The ELBO (the objective at the right side of Eq.~\ref{eq:elbo}) lower bounds the log-likelihood of observed data, and the gap vanishes at the global optimum. Often, the density forms of $p(\vz)$ and $q_\phi(\vz|\vx)$ are chosen such that their KL-divergence can be written analytically in a closed-form expression (e.g., $p(\vz)$ is $N(0,I)$ and $q_\phi(\vz|\vx)$ is $N(\mu_\phi(\vx), \Sigma_\phi(\vx))$) \citep{kingma2013auto}. 
In such cases, the ELBO can be efficiently optimized (to a stationary point) using stochastic first order methods where both expectations are estimated using mini-batches. Further, in cases when $q_\phi(\cdot)$ can be written as a continuous transformation of a fixed base distribution (e.g., the standard normal distribution), a low variance estimate of the gradient over $\phi$ can be obtained by coordinate transformation (also referred as reparametrization) \citep{fu2006gradient,kingma2013auto,rezende2014stochastic}.   

%generative model starts with disentangled prior; 
\subsection{Generative story: disentangled prior}
Most VAE based generative models for real datasets (e.g., text, images, etc.) already work with a relatively simple and disentangled prior $p(\vz)$ having no interaction among the latent dimensions (e.g., the standard Gaussian $N(0,I)$) \citep{bowman2015generating,miao2016neural,hou2017deep,zhao2017towards}. The complexity of the observed data is absorbed in the conditional distribution $p_\theta(\vx|\vz)$ which encodes the interactions among the latents. Hence, as far as the generative modeling is concerned, disentangled prior sets us in the right direction. 

%define approximate prior - this is what we want to disentangle so that inferred latents are disentangled; KL inequality - why ideally VAE should give disentangled latents and to extent it does also, however limitation of opt -> add it as a regularizer
\subsection{Inferring disentangled latents}
Although the generative model starts with a disentangled prior, our main objective is to \emph{infer} disentangled latents which are potentially conducive for various goals mentioned in Sec.~\ref{sec:intro} (e.g., invariance, transferability, interpretability). To this end, we consider the density over the inferred latents induced by the approximate posterior inference mechanism, 
\begin{align}
q_\phi(\vz) = \int q_\phi(\vz|\vx) p(\vx) d\vx,
\end{align}
which we will subsequently refer to as the \emph{inferred prior} or \emph{expected variational posterior} ($p(\vx)$ is the true data distribution that we have only samples from). For inferring disentangled factors, this should be factorizable along the dimensions, i.e., $q_\phi(\vz)=\prod_{i} q_i(z_i)$, or equivalently $q_{i|j}(z_i|z_j)=q_i(z_i),\,\forall\, i,j$. %\footnote{In some scenarios, dependent generative factors cannot be ruled out (e.g., gender and beard}. 
This can be achieved by minimizing a suitable distance between the inferred prior $q_\phi(\vz)$ and the disentangled generative prior $p(\vz)$. We can also define \emph{expected posterior} as $p_\theta(\vz) = \int p_\theta(\vz|\vx)p(\vx)d\vx$. If we take KL-divergence as our choice of distance, by relying on its pairwise convexity (i.e., $\kl(\lambda p_1+(1-\lambda) p_2\Vert \lambda q_1 + (1-\lambda) q_2) \leq \lambda \kl(p_1\Vert q_1) + (1-\lambda) \kl(p_2\Vert q_2)$) \citep{van2014renyi}, we can show that the distance between $q_\phi(\vz)$ and $p_\theta(\vz)$ is bounded by the objective of the variational inference: % (the ELBO in Eq.~\eqref{eq:elbo}):
\begin{align}
\kl (q_\phi(\vz)\Vert p_\theta(\vz)) = \kl(\expect_{\vx\sim p(\vx)} q_\phi(\vz|\vx)\Vert \expect_{\vx\sim p(\vx)} p_\theta(\vz|\vx)) \leq \expect_{\vx\sim p(\vx)}\, \kl (q_\phi(\vz|\vx)\Vert p_\theta(\vz|\vx)).
\label{eq:ub}
\end{align}
%\blue{consider including a lemma with proof?}\\
%\begin{lem}
%xxx 
%\end{lem}
In general, the prior $p(\vz)$ and expected posterior $p_\theta(\vz)$ will be different, although they may be close (they will be same when $p_\theta(\vx)=\int p_\theta(\vx|\vz)p(\vz)d\vz$ is equal to $p(\vx)$). 
%For prior-posterior pairs that are \emph{Bayes plausible} \citep{kamenica2011bayesian} expected posterior  
Hence, variational posterior inference of latent variables with disentangled prior naturally encourages inferring factors that are \emph{close} to being disentangled. We think this is the reason that the original VAE (Eq. \eqref{eq:elbo}) has also been observed to exhibit some disentangling behavior on simple datasets such as MNIST \citep{kingma2013auto}. However, this behavior does not carry over to more complex datasets \citep{aubry2014seeing,liu2015deep,higgins2016beta}, unless extra supervision on the generative factors is provided \citep{kulkarni2015deep,karaletsos2015bayesian}. This can be due to: (i) $p(\vx)$ and $p_\theta(\vx)$ being far apart which in turn causes $p(\vz)$ and $p_\theta(\vz)$ being far apart, and (ii) the non-convexity of the ELBO objective which prevents us from achieving the global minimum of $\expect_\vx \kl(q_\phi(\vz|\vx)\Vert p_\theta(\vz|\vx))$ (which is $0$ and implies $\kl(q_\phi(\vz)\Vert p_\theta(\vz))=0$). 
In other words, maximizing the ELBO (Eq.~\eqref{eq:elbo}) might also result in reducing the value of $\kl(q_\phi(\vz)\Vert p(\vz))$, however, due to the aforementioned reasons, the gap between $\kl(q_\phi(\vz)\Vert p(\vz))$ and $\expect_\vx\, \kl (q_\phi(\vz|\vx)\Vert p_\theta(\vz|\vx))$ could be large at the stationary point of convergence. Hence, minimizing $\kl(q_\phi(\vz)\Vert p(\vz))$ or any other suitable distance $D(q_\phi(\vz), p(\vz))$ explicitly will give us better control on the disentanglement. 
This motivates us to add $D(q_\phi(\vz)\Vert p(\vz))$ as part of the objective to encourage disentanglement during inference, i.e.,
\begin{align}
\max_{\theta,\phi}\, \expect_\vx \left[\expect_{\vz\sim q_\phi(\vz|\vx)}\left[ \log p_\theta(\vx|\vz)\right]  - \kl (q_\phi(\vz|\vx)\Vert p(\vz))\right] - \lambda\, D(q_\phi(\vz)\Vert p(\vz)),
\label{eq:regelbo}
\end{align}
where $\lambda$ controls its contribution to the overall objective. We refer to this as DIP-VAE (for Disentangled Inferred Prior) subsequently. 

%optimizing KL is not tractable directly; implicit optimization possible (GAN framework); we resort to a simpler alternative - matching the second moments which would mean uncorrelated z's 
Optimizing \eqref{eq:regelbo} directly is not tractable if $D(\cdot,\cdot)$ is taken to be the KL-divergence  $\kl(q_\phi(\vz)\Vert p(\vz))$, which does not have a closed-form expression. One possibility is use the variational formulation of the KL-divergence \citep{nguyen2010estimating,nowozin2016f} that needs only samples from $q_\phi(\vz)$ and $p(\vz)$ to estimate a lower bound to $\kl(q_\phi(\vz)\Vert p(\vz))$. However, this would involve optimizing for a third set of parameters $\psi$ for the KL-divergence estimator, and would also change the optimization to a saddle-point (min-max) problem which has its own optimization challenges (e.g., gradient vanishing as encountered in training generative adversarial networks with KL or Jensen-Shannon (JS) divergences \citep{goodfellow2014generative,arjovsky2017towards}). Taking $D$ to be another suitable distance between $q_\phi(\vz)$ and $p(\vz)$ (e.g., integral probability metrics like Wasserstein distance \citep{sriperumbudur2009integral}) might alleviate some of these issues \citep{arjovsky2017wasserstein} but will still involve complicating the optimization to a saddle point problem in three set of parameters\footnote{Nonparametric distances like maximum mean discrepancy (MMD) with a characteristic kernel \citep{gretton2012kernel} is also an option, however it has its own challenges when combined with stochastic optimization \citep{dziugaite2015training,li2015generative}.}. It should also be noted that using these variational forms of the distances will still leave us with an approximation to the actual distance.  %Even if we ignore the training instabilities involved with the non-convex saddle point problems, using variational formulation for KL-divergence would still leave us with an approximation of the true objective at best. 

We adopt a simpler yet effective alternative of matching the moments of the two distributions. Matching the covariance of the two distributions will amount to decorrelating the dimensions of $\vz\sim q_\phi(\vz)$ if $p(\vz)$ is $N(0,I)$. Let us denote %$\vmu_{q(\vz)} := \expect_{q(\vz)} \vz$, and 
$\text{Cov}_{q(\vz)}[\vz] := \expect_{q(\vz)} \left[(\vz-\expect_{q(\vz)}[\vz])(\vz-\expect_{q[\vz]}(\vz))^\top\right]$. By the law of total covariance, the covariance of $\vz\sim q_\phi(\vz)$ is given by %\blue{cite a book here}
\begin{align}
%\expect_{q_\theta(\vz)} \left[(\vz-\expect_{q_\theta(\vz)}\vz)(\vz-\expect_{q_\theta(\vz)}\vz)^\top\right] = \expect_{p(\vx)} 
\text{Cov}_{q_\phi(\vz)}[\vz] = \expect_{p(\vx)} \text{Cov}_{q_\phi(\vz|\vx)}[\vz] + \text{Cov}_{p(\vx)} \left(\expect_{q_\phi(\vz|\vx)} [\vz]\right),
\label{eq:totalcov}
\end{align}
where $\expect_{q_\phi(\vz|\vx)}[\vz]$ and $\text{Cov}_{q_\phi(\vz|\vx)}[\vz]$ are random variables that are functions of the random variable $\vx$ ($\vz$ is marginalized over). Most existing work on the VAE models uses $q_\phi(\vz|\vx)$ having the form $N(\vmu_\phi(\vx), \vSigma_\phi(\vx))$, where $\vmu_\phi(\vx)$ and $\vSigma_\phi(\vx)$ are the outputs of a deep neural net parameterized by $\phi$. In this case Eq.~\eqref{eq:totalcov} reduces to $\text{Cov}_{q_\phi(\vz)}[\vz] = \expect_{p(\vx)} [\vSigma_\phi(\vx)] + \text{Cov}_{p(\vx)}[\vmu_\phi(\vx)]$, which we want to be close to the Identity matrix.
For simplicity, we choose entry-wise squared $\ell_2$-norm as the measure of proximity.
Further, $\vSigma_\phi(\vx)$ is commonly taken to be a diagonal matrix which means that cross-correlations (off-diagonals) between the latents are due to only $\text{Cov}_{p(\vx)}[\vmu_\phi(\vx)]$. This suggests two possible options for the disentangling regularizer: (i) regularizing only $\text{Cov}_{p(\vx)}[\vmu_\phi(\vx)]$ which we refer as DIP-VAE-I, (ii) regularizing $\text{Cov}_{q_\phi(\vz)}[\vz]$ which we refer as DIP-VAE-II. Penalizing just the off-diagonals in both cases will lead to lowering the diagonal entries of $\text{Cov}_{p(\vx)}[\vmu_\phi(\vx)]$ as the $ij$'th off-diagonal is really a derived attribute obtained by multiplying the square-roots of $i$'th and $j$'th diagonals (for each example $\vx\sim p(\vx)$, followed by averaging over all examples). This can be compensated in DIP-VAE-I by a regularizer on the diagonal entries of $\text{Cov}_{p(\vx)}[\vmu_\phi(\vx)]$ which pulls these towards $1$. We opt for two separate hyperparameters controlling the relative importance of the loss on the diagonal and off-diagonal entries as follows:
\begin{align}
\max_{\theta,\phi}\, \text{ELBO}(\theta,\phi) - \lambda_{od}\,\sum_{i\neq j}\left[\text{Cov}_{p(\vx)}[\vmu_\phi(\vx)]\right]^2_{ij} - \lambda_{d}\,\sum_{i}\left(\left[\text{Cov}_{p(\vx)}[\vmu_\phi(\vx)]\right]_{ii}-1\right)^2.
\label{eq:regcovelbo}
\end{align}
The regularization terms involving $\text{Cov}_{p(\vx)}[\vmu_\phi(\vx)]$ in the above objective \eqref{eq:regcovelbo} can be efficiently optimized using SGD, where $\text{Cov}_{p(\vx)}[\vmu_\phi(\vx)]$ can be estimated using the current minibatch\footnote{We also tried an alternative of maintaining a running estimate of $\text{Cov}_{p(\vx)}[\vmu_\phi(\vx)]$ which is updated with every minibatch of $\vx\sim p(\vx)$, however we did not observe a significant improvement over the simpler approach of estimating these using only current minibatch.}.
%We maintain a running estimate of $\text{Cov}_{p(\vx)}[\vmu_\phi(\vx)]$ which is updated with every minibatch of $\vx\sim p(\vx)$. The gradient for the current minibatch can be computed by treating the previous estimate of $\text{Cov}_{p(\vx)}[\vmu_\phi(\vx)]$ as constant. 

For DIP-VAE-II, we have the following optimization problem:
\begin{align}
\max_{\theta,\phi}\, \text{ELBO}(\theta,\phi) - \lambda_{od}\,\sum_{i\neq j}\left[\text{Cov}_{q_\phi(\vz)}[\vz]\right]^2_{ij} - \lambda_{d}\,\sum_{i}\left(\left[\text{Cov}_{q_\phi(\vz)}[\vz]\right]_{ii}-1\right)^2.
\label{eq:regcovelbo2}
\end{align}
As discussed earlier, the term $\expect_{p(\vx)} \text{Cov}_{q_\phi(\vz|\vx)}[\vz]$ contributes only to the diagonals of $\text{Cov}_{q_\phi(\vz)}[\vz]$. Penalizing the off-diagonals of $\text{Cov}_{p(\vx)}[\vmu_\phi(\vx)]$ in the Objective~\eqref{eq:regcovelbo2} will contribute to reduction in the magnitude of its diagonals as discussed earlier. As the regularizer on the diagonals is not directly on $\text{Cov}_{p(\vx)}[\vmu_\phi(\vx)]$, unlike DIP-VAE-I, it will be not be able to keep $[\text{Cov}_{p(\vx)}[\vmu_\phi(\vx)]]_{ii}$ close to $1$: the reduction in $[\text{Cov}_{p(\vx)}[\vmu_\phi(\vx)]]_{ii}$ will be accompanied by increase in $[\expect_{p(\vx)} \vSigma_\phi(\vx)]_{ii}$ such that their sum remains close to $1$. In datasets where the number of generative factors is less than the latent dimension, DIP-VAE-II is more suitable than DIP-VAE-I as keeping all dimensions \emph{active} might result in \emph{splitting} of an attribute across multiple dimensions, hurting the goal of disentanglement. 

It is also possible to match higher order central moments of $q_\phi(\vz)$ and the prior $p(\vz)$. In particular, third order central moments (and moments) of the zero mean Gaussian prior are zero, hence $\ell_2$ norm of third order central moments of $q_\phi(\vz)$ can be penalized. 

%  However, as the entanglement is mainly reflected in the off-diagonal entries of this matrix, we opt for two separate hyperparameters controlling the relative importance of the loss on the diagonal and off-diagonal entries. This gives rise to the following optimization problem for inference:
%comparison with beta-vae objective, particular in the case of diagonal posterior

\subsection{Comparison with $\beta$-VAE}
Recently proposed $\beta$-VAE \citep{higgins2016beta} proposes to modify the ELBO by upweighting the $\kl(q_\phi(\vz|\vx)\Vert p(\vz))$ term in order to encourage the inference of disentangled factors:
\begin{align}
\max_{\theta,\phi}\, \expect_\vx \left[\expect_{\vz\sim q_\phi(\vz|\vx)}\left[ \log p_\theta(\vx|\vz)\right]  - \beta\, \kl (q_\phi(\vz|\vx)\Vert p(\vz))\right],
\label{eq:betavae}
\end{align}
where $\beta$ is taken to be great than $1$. Higher $\beta$ is argued to encourage disentanglement at the cost of reconstruction error (the likelihood term in the ELBO). Authors report empirical results with $\beta$ ranging from $4$ to $250$ depending on the dataset. 
As already mentioned, most VAE models proposed in the literature, including $\beta$-VAE, work with $N(\mbf{0},\vI)$ as the prior $p(\vz)$ and $N(\vmu_\phi(\vx),\vSigma_\phi(\vx))$ with diagonal $\vSigma_\phi(\vx)$ as the approximate posterior $q_\phi(\vz|\vx)$. This reduces the objective \eqref{eq:betavae} to
\begin{align}
\max_{\theta,\phi}\, \expect_\vx \left[\expect_{\vz\sim q_\phi(\vz|\vx)}\left[ \log p_\theta(\vx|\vz)\right]  - \frac{\beta}{2} \left( \sum_i\left( \left[\vSigma_\phi(\vx)\right]_{ii} - \ln\left[\vSigma_\phi(\vx)\right]_{ii}\right) + \lVert \vmu_\phi(\vx)\rVert_2^2 \right) \right].
\label{eq:betavaered}
\end{align}
For high values of $\beta$, $\beta$-VAE would try to pull $\vmu_\phi(\vx)$ towards zero and $\vSigma_\phi(\vx)$ towards the identity matrix (as the minimum of $x-\ln x$ for $x>0$ is at $x=1$), thus making the approximate posterior $q_\phi(\vz|\vx)$ insensitive to the observations. This is also reflected in the quality of the reconstructed samples which is worse than VAE ($\beta=1$), particularly for high values of $\beta$. 
Our proposed method does not have such increased tension between the likelihood term and the disentanglement objective, and the sample quality with our method is on par with the VAE. 

Finally, we note that both $\beta$-VAE and our proposed method encourage disentanglement of inferred factors by pulling $\text{Cov}_{q_\phi(\vz)}(\vz)$ in Eq.~\eqref{eq:totalcov} towards the identity matrix: $\beta$-VAE attempts to do it by making $\text{Cov}_{q_\phi(\vz|\vx)}(\vz)$ close to $\vI$ and $\expect_{q_\phi(\vz|\vx)}(\vz)$ close to $\mbf{0}$ individually for all observations $\vx$, while the proposed method directly works on $\text{Cov}_{q_\phi(\vz)}(\vz)$ (marginalizing over the observations $\vx$) which retains the sensitivity of $q_\phi(\vz|\vx)$ to the conditioned-upon observation.

%Gaussian case?
\section{Quantifying disentanglement: SAP Score}
\label{sec:sapscore}
\citet{higgins2016beta} propose a metric to evaluate the disentanglement performance of the inference mechanism, assuming that the ground truth generative factors are available. It works by first sampling a generative factor $y$, followed by sampling $L$ pairs of examples such that for each pair, the sampled generative factor takes the same value. Given the inferred $\vz_x := \vmu_\phi(\vx)$ for each example $\vx$, they compute the absolute difference of these vectors for each pair, followed by averaging these difference vectors. This average difference vector is assigned the label of $y$. By sampling $n$ such minibatches of $L$ pairs, we get $n$ such averaged difference vectors for the factor $y$. This process is repeated for all generative factors. A low capacity multiclass classifier is then trained on these vectors to predict the identities of the corresponding generative factors. Accuracy of this classifier on the difference vectors for test set is taken to be a measure of disentanglement. We evaluate the proposed method on this metric and refer to this as {\bf Z-diff score} subsequently. 

We observe in our experiments that the Z-diff score  \citep{higgins2016beta} is not correlated well with the qualitative disentanglement at the decoder's output as seen in the latent traversal plots (obtained by varying only one latent while keeping the other latents fixed). It also depends on the multiclass classifier used to obtain the score. We propose a new metric, referred as {\bf Separated Attribute Predictability (SAP) score}, that is better aligned with the qualitative disentanglement observed in the latent traversals and also does not involve training any classifier. It is computed as follows: {\bf (i)} We first construct a $d\times k$ score matrix $S$ (for $d$ latents and $k$ generative factors) whose $ij$'th entry is the linear regression or classification score (depending on the generative factor type) of predicting $j$'th factor using only $i$'th latent $[\vmu_\phi(\vx)]_i$. For regression, we take this to be the $R^2$ score obtained with fitting a line (slope and intercept) that minimizes the linear regression error (for the test examples). The $R^2$ score is given by $\left(\frac{\text{Cov}([\vmu_\phi(\vx)]_i, \vy_j)}{\sigma_{[\vmu_\phi(\vx)]_i}\sigma_{\vy_j}}\right)^2$ and ranges from $0$ to $1$, with a score of $1$ indicating that a linear function of the $i$'th inferred latent explains all variability in the $j$'th generative factor. For classification, we fit one or more thresholds (real numbers) directly on $i$'th inferred latents for the test examples that minimize the balanced classification errors, and take $S_ij$ to be the balanced classification accuracy of the $j$'th generative factor. For inactive latent dimensions (having  $\sigma_{[\vmu_\phi(\vx)]_i} = [\text{Cov}_{p(x)} [\vmu_\phi(\vx)]]_{ii}$ close to $0$), we take $S_{ij}$ to be $0$. 
{\bf (ii)} For each column of the score matrix $S$ which corresponds to a generative factor, we take the difference of top two entries (corresponding to top two most predictive latent dimensions), and then take the mean of these differences as the final {\bf SAP score}. Considering just the top scoring latent dimension for each generative factor is not enough as it does not rule out the possibility of the factor being captured by other latents. A high SAP score indicates that each generative factor is primarily captured in only one latent dimension. Note that a high SAP score does not rule out one latent dimension capturing two or more generative factors well, however in many cases this would be due to the generative factors themselves being correlated with each other, which can be verified empirically using ground truth values of the generative factors (when available). Further, a low SAP score does not rule out good disentanglement in cases when two (or more) latent dimensions might be correlated strongly with the same generative factor and poorly with other generative factors. 
The generated examples using single latent traversals may not be realistic for such models, and DIP-VAE discourages this from happening by enforcing decorrelation of the latents. 
However, the SAP score computation can be adapted to such cases by  grouping the latent dimensions based on correlations and getting the score matrix at group level, which can be fed as input to the second step to get the final SAP score.

%% file: experiments.tex
\section{Experiments}
\begin{table}[t]
\centering
\caption{Z-diff score \cite{higgins2016beta}, the proposed SAP score  and reconstruction error (per pixel) on the test sets for 2D Shapes and CelebA ($\beta_1=4,\beta_2=60, \lambda=10, \lambda_1=5, \lambda_2=500$ for 2D Shapes; $\beta_1=4,\beta_2=32, \lambda=2, \lambda_1=1, \lambda_2=80$ for CelebA). For the results on a wider range of hyperparameter values, refer to Fig.~\ref{fig:recon-zdiff-dsprites} and Fig.~\ref{fig:recon-zdiff-celeba}.}
\begin{tabular}{||c c c c| c c c||} 
 \hline
 \multirow{2}{*}{Method} & \multicolumn{3}{c|}{2D Shapes} & \multicolumn{3}{c||}{CelebA} \\

 & Z-diff & SAP & Reconst. error & Z-diff & SAP & Reconst. error \\
 \hline\hline
 VAE & 81.3 & 0.0417 & 0.0017 & 7.5 & 0.35 &   0.0876 \\ 
 $\beta$-VAE ($\beta$=$\beta_1$) & 80.7 & 0.0811 &  0.0032  & 8.1 & 0.48 & 0.0937 \\
 $\beta$-VAE ($\beta$=$\beta_2$) & 95.7 & 0.5503 & 0.0113  & 6.4 & 3.72 & 0.1572 \\
 %DIP-VAE-I & \textbf{98.7} & 0.1889 & 0.0018  &  {\bf 11.31 } & 1.94 & 0.0911 \\
 DIP-VAE-I ($\lambda_{od}=\lambda$) & 98.7 & 0.1889 & 0.0018  & 14.8  & 3.69 & 0.0904 \\
 DIP-VAE-II ($\lambda_{od}=\lambda_1)$& 95.3 & 0.2188 & 0.0023  & 7.1  & 2.94 & 0.0884 \\
 DIP-VAE-II ($\lambda_{od}=\lambda_2)$& 98.0 & 0.5253 &0.0079   & 11.5  & 3.93 & 0.1477 \\
 \hline
\end{tabular}
%\vspace{-10mm}
\label{tab:zdiff}
\end{table}

\begin{figure}[t]
    \centering
    \includegraphics[scale=0.4]{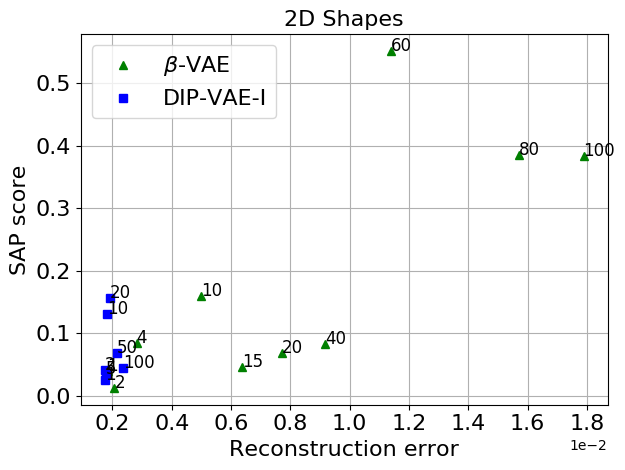} \includegraphics[scale=0.4]{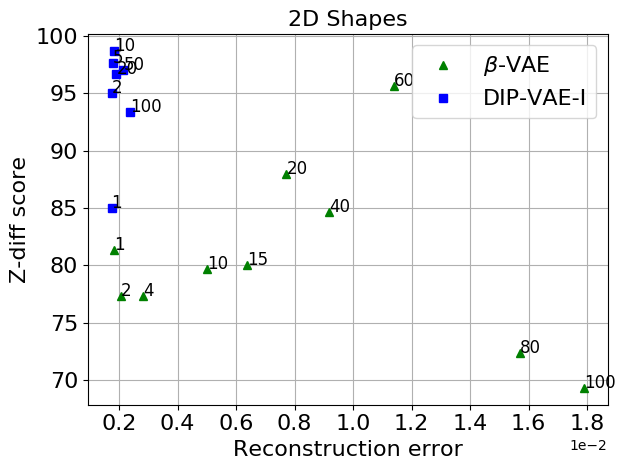}
        \includegraphics[scale=0.4]{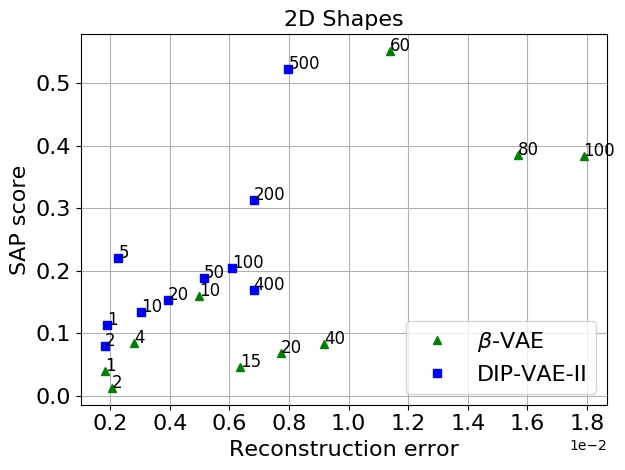} \includegraphics[scale=0.4]{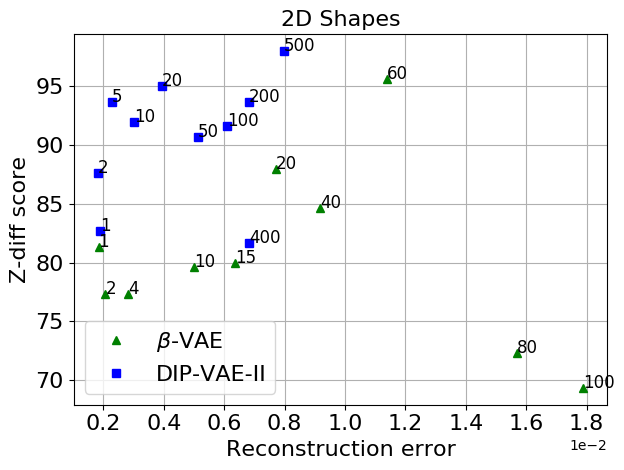}
\caption{Proposed Separated Atomic Predictability (SAP) score and the Z-diff disentanglement score \citep{higgins2016beta} as a function of average reconstruction error (per pixel) on the test set of 2D Shapes data for $\beta$-VAE and the proposed DIP-VAE. The plots are generated by varying $\beta$ for $\beta$-VAE, and $\lambda_{od}$ for DIP-VAE-I and DIP-VAE-II (the number next to each point is the value of these hyperparameters, respectively).}
%\caption{Disentanglement metric score \citep{higgins2016beta} as a function of average reconstruction error (per pixel) on the test set for $\beta$-VAE and the proposed DIP-VAE (\emph{left:} CelebA, \emph{right:} 2D Shapes). The plots are generated by varying $\beta$ for $\beta$-VAE, and $\lambda_{od}$ for DIP-VAE with $\lambda_{d}$ set to $10\lambda_{od}$. The number next to each point in the plots is the value of $\beta$ ($\beta$-VAE) or $\lambda_{od}$ (proposed DIP-VAE).}
\label{fig:recon-zdiff-dsprites}
\end{figure}

We evaluate our proposed method, DIP-VAE, on three datasets -- {(i) \bf CelebA } \citep{liu2015deep}: It consists of $202,599$ RGB face images of celebrities. We use $64\times 64\times 3$ cropped images as used in several earlier works, using $90\%$ for training and $10\%$ for test. {(ii) \bf 3D Chairs} \citep{aubry2014seeing}: It consists of 1393 chair CAD models, with each model rendered from 31 azimuth angles and 2 elevation
angles. Following earlier work \citep{yang2015weakly,dosovitskiy2015learning} that ignores near-duplicates, we use a subset of 809 chair models in our experiments. We use the binary masks of the chairs as the observed data in our experiments following \citep{higgins2016beta}. First $80\%$ of the models are used for training and the rest are used for test. 
{\bf (iii) 2D Shapes} \citep{dsprites17}: This is a synthetic dataset of binary 2D shapes generated from the Cartesian product of the shape (heart, oval and square), $x$-position  (32 values), $y$-position  (32 values), scale (6 values) and rotation (40 values). We consider two baselines for the task of unsupervised inference of disentangled factors: (i) VAE \citep{kingma2013auto,rezende2014stochastic}, and (ii) the recently proposed $\beta$-VAE \citep{higgins2016beta}. To be consistent with the evaluations in \citep{higgins2016beta}, we use the same CNN network architectures (for our encoder and decoder), and same latent dimensions as used in \citep{higgins2016beta} for CelebA, 3D Chairs, 2D Shapes datasets.

{\bf Hyperparameters.~} For the proposed DIP-VAE-I, in all our experiments we vary $\lambda_{od}$ in the set $\{1, 2, 5, 10, 20, 50,100,500\}$ while fixing $\lambda_{d}=10\lambda_{od}$ for 2D Shapes and 3D Chairs, and $\lambda_{d}=50\lambda_{od}$ for CelebA. For DIP-VAE-II, we fix $\lambda_{od}=\lambda_d$ for 2D Shapes, and $\lambda_{od}=2\lambda_{d}$ for CelebA. Additionally, for DIP-VAE-II we also penalize the $\ell_2$-norm of third order central moments of $q_\phi(\vz)$ with hyperparameter $\lambda_3=200$ for 2D Shapes data ($\lambda_3=0$ for CelebA). For $\beta$-VAE, we experiment with $\beta=\{1,2,4,8,16,25,32,64,100,128$
$,200,256\}$ (where $\beta=1$ corresponds to the VAE). We used a batch size of $400$ for all 2D Shapes experiments and $100$ for all CelebA experiments. For both CelebA and 2D Shapes, we show the results %for the best performing models 
in terms of the Z-diff score \cite{higgins2016beta}, the proposed SAP score, and reconstruction error. %as the hyperparameter  which was introduced in \citep{higgins2016beta}. 
For 3D Chairs data, only two ground truth generative factors are available and the quantitative scores for these are saturated near the peak values, hence we show only the latent traversal plots which we based on our subjective evaluation of the reconstruction quality and disentanglement (shown in Appendix). 

\begin{figure}[t]
    \centering
    \includegraphics[scale=0.4]{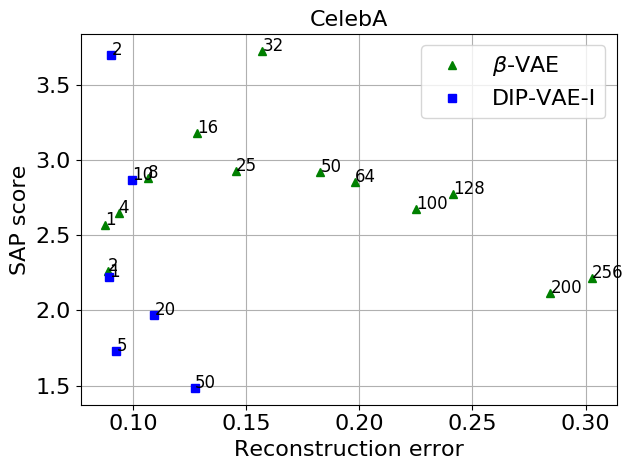} \includegraphics[scale=0.4]{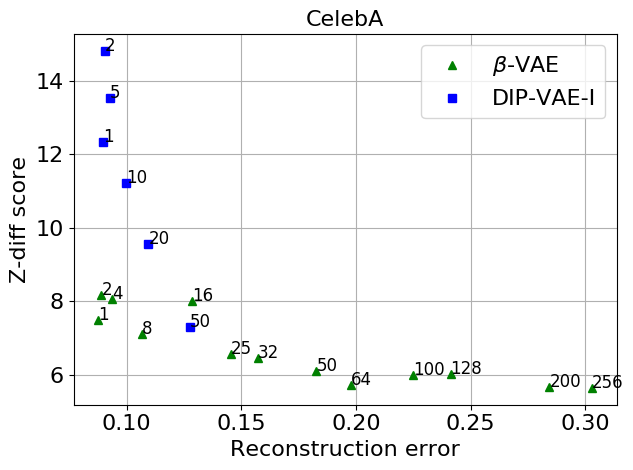}
        \includegraphics[scale=0.4]{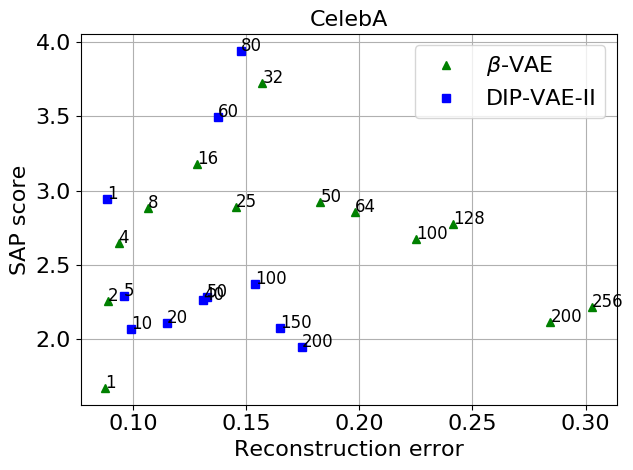} \includegraphics[scale=0.4]{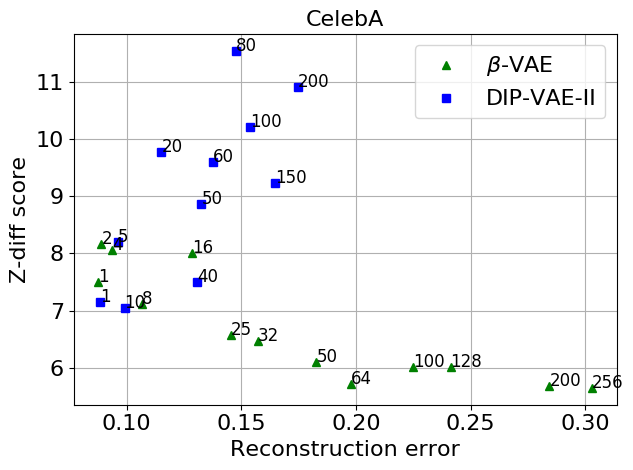}
\caption{The proposed SAP score and the Z-diff score \citep{higgins2016beta} as a function of average reconstruction error (per pixel) on the test set of CelebA data for $\beta$-VAE and the proposed DIP-VAE. The plots are generated by varying $\beta$ for $\beta$-VAE, and $\lambda_{od}$ for DIP-VAE-I and DIP-VAE-II (the number next to each point is the value of these hyperparameters, respectively).}
\label{fig:recon-zdiff-celeba}
\end{figure}

% Dsprites disentangle figures
\begin{figure}
\centering
\hspace{-7mm}\includegraphics[width=1.05\textwidth]{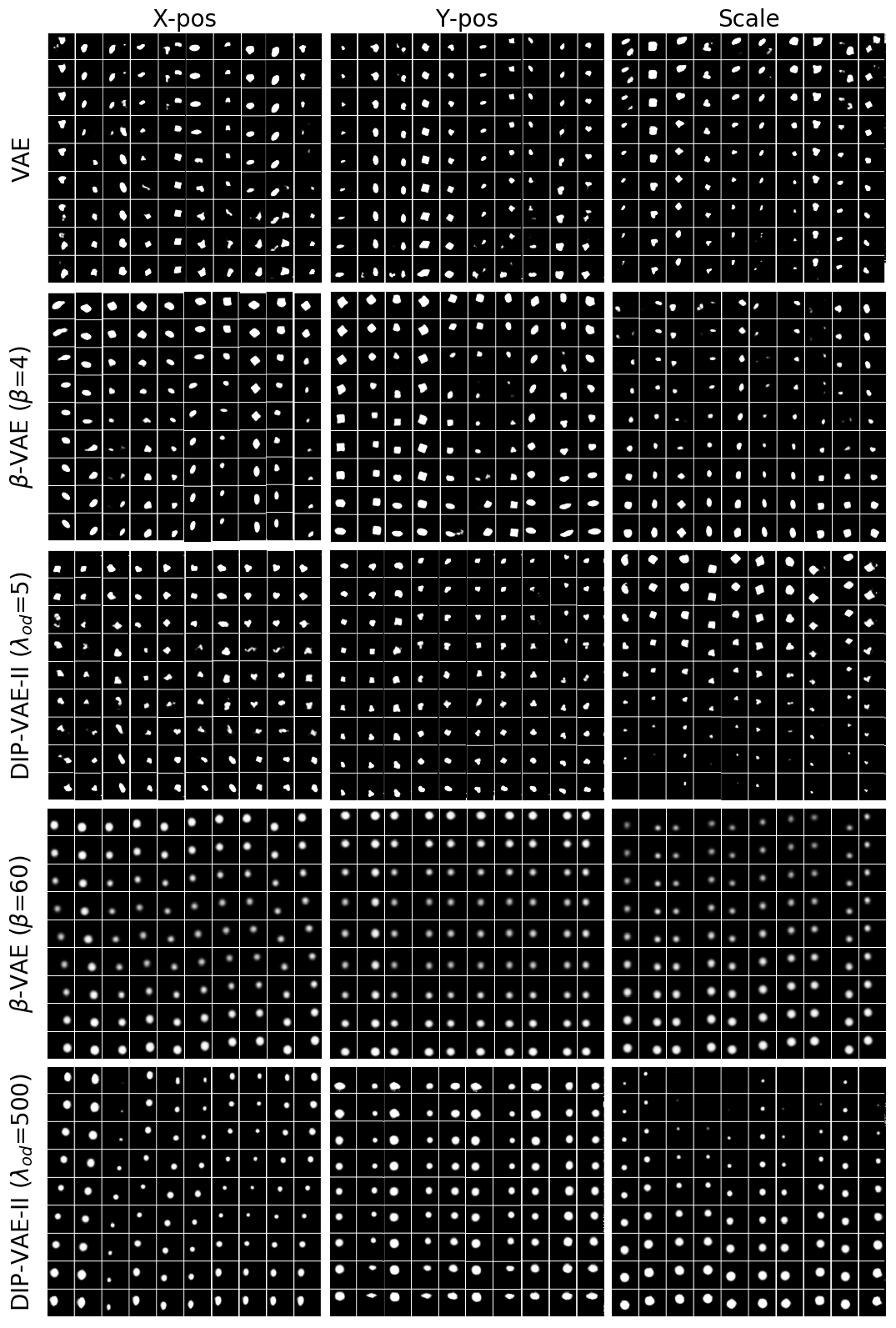}
\caption{Qualitative results for disentanglement in 2D Shapes dataset \citep{dsprites17}. SAP scores, Z-diff scores and reconstruction errors for the methods (rows) can be read from Fig.~\ref{fig:recon-zdiff-dsprites}. }
\label{fig:sweeps_dsprites}
\end{figure}

\begin{figure}[h]
\centering
\hspace{-7mm}\includegraphics[width=1.05\textwidth]{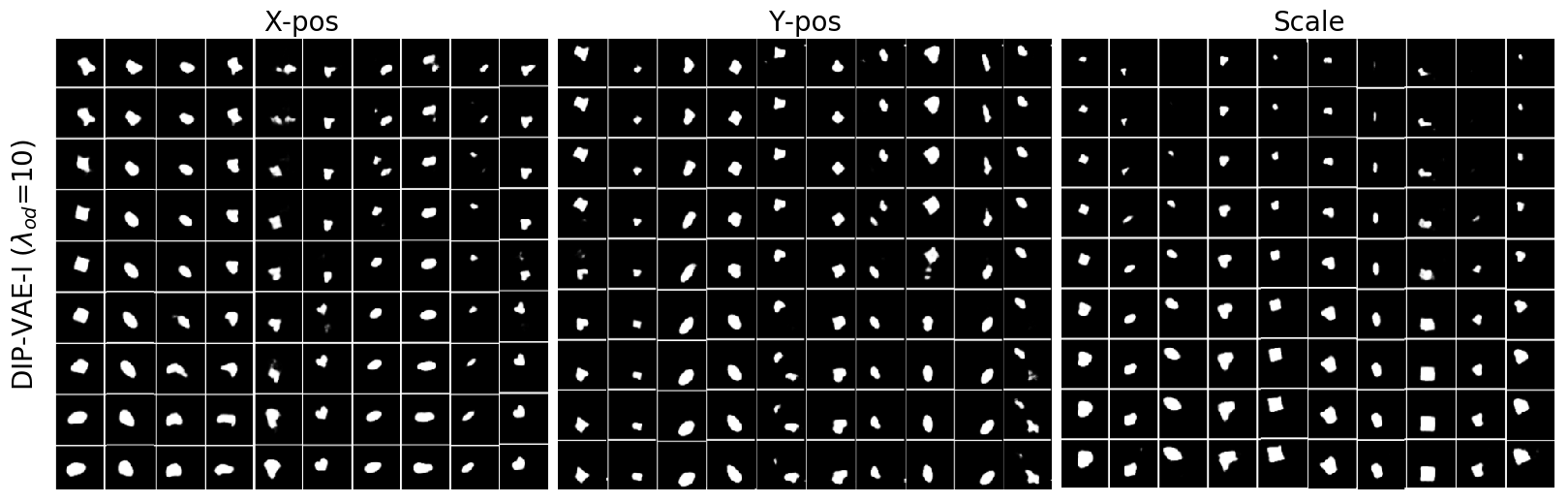}
\caption{Qualitative results for disentanglement in 2D Shapes dataset \citep{dsprites17} for DIP-VAE-I (SAP score $0.1889$).}
\label{fig:sweeps_dsprites_dipvaeI}
\end{figure}

%\begin{figure}[h]
%\centering
%\includegraphics[width=\textwidth]{figures/z_cov_beta-vae.png}
%\includegraphics[width=\textwidth]{figures/z_cov_dipvae2.png}
%\includegraphics[width=\textwidth]{figures/z_corr_rev.png}
%\includegraphics[width=\textwidth]{figures/label_corr_rev.png}
%\caption{2D Shapes data \citep{higgins2016beta}: Covariance between inferred latents for $\beta$-VAE \emph{(top)} and DIP-VAE-II \emph{(bottom)}.}
%\label{fig:corrs-shapes}
%\end{figure}

\begin{table}[t]
	\centering
	\caption{Attribute classification accuracy on CelebA: A classifier $\vw^k=\frac{1}{|\vx_i:y_i^k=1|}\sum_{\vx_i:y_i^k=1}\mu_\phi(\vx_i)-\frac{1}{|\vx_i:y_i^k=0|}\sum_{\vx_i:y_i^k=0}\mu_\phi(\vx_i)$ is computed for every attribute $k$ using the training set and a bias is learned by minimizing the hinge loss. Accuracy on other attributes stays about same across all methods.}
    
\begin{tabu} to \textwidth {||c X[c] X[c] X[c] X[c] X[c] X[c] X[c] X[c] X[c] X[c] X[c] X[c]||}  
		\hline
		\rot{Method} & \rot{Arched Eyebrows} & \rot{Attractive} & \rot{Bangs} & \rot{Black hair}& \rot{Blond hair} & \rot{Heavy makeup}& \rot{Male}& \rot{Mouth slighly open} & \rot{No Beard}& \rot{Wavy hair} & \rot{Wearing hat} & \rot{Wearing lipstick} \\ [0.1ex] 
		\hline\hline
		VAE & 71.8 & 73.0& 89.8& 78.0& 88.9 &  79.6& 83.9&{\bf 76.3} & {\bf 87.3} & 70.2 & 95.8 & 83.0 \\
		$\beta=2$ & 71.6 & 72.6 & 90.6 & 79.3 & 89.1 &   79.3 & 83.5 & 76.1 & 86.9 & 67.8 & 95.9 & 82.4 \\
		$\beta=4$ & 71.6 & 72.6 & 90.0 & 76.6 & 88.9 & 77.8 & 82.3 & 75.7 & 85.3 & 66.8 & 95.8 & 80.6 \\ 
		$\beta=8$ & 71.6 & 71.7 & 90.0 & 76.0 & 87.2 &  76.2 & 80.5 & 73.1 & 85.3 & 63.7 & 95.8 & 79.6 \\
		DIP-VAE-I & {\bf 73.7} & {\bf 73.2}& {\bf 90.9}& {\bf 80.6} & {\bf 91.9} &{\bf 81.5}& {\bf 85.9} & 75.9 &85.3 & {\bf 71.5} & {\bf 96.2} & {\bf 84.7} \\
		\hline
        \end{tabu}
	
	\label{tab:binclass}
	%\vspace{-5mm}
\end{table}

{\bf Disentanglement scores and reconstruction error.~ }
For the Z-diff score \citep{higgins2016beta}, in all our experiments we use a one-vs-rest linear SVM with weight on the hinge loss $C$ set to $0.01$ and weight on the regularizer set to $1$. %\citet{higgins2016beta} argue that this metric captures the disentangled property of the inferred latents reasonably well. 
Table~\ref{tab:zdiff} shows the Z-diff scores and the proposed SAP scores along with reconstruction error (which directly corresponds to the data likelihood) for the test sets of CelebA and 2D Shapes data. 
Further we also show the plots of how the Z-diff score and the proposed SAP score change with the reconstruction error as we vary the hyperparameter for both methods ($\beta$ and $\lambda_{od}$, respectively) in Fig.~\ref{fig:recon-zdiff-dsprites} (for 2D Shapes data) and Fig.~\ref{fig:recon-zdiff-celeba} (for CelebA data).  The proposed DIP-VAE-I gives much higher Z-diff score at little to no cost on the reconstruction error when compared with VAE ($\beta=1$) and $\beta$-VAE, for both 2D Shapes and CelebA datasets. 
%The proposed DIP-VAE-I outperforms $\beta$-VAE in terms of the Z-diff score and reconstruction error for both 2D Shapes and CelebA. 
However, we observe in the decoder's output for single latent traversals (varying a single latent while keeping others fixed, shown in Fig.~\ref{fig:sweeps_dsprites} and Fig.~\ref{fig:sweeps_dsprites_dipvaeI}) that a high Z-diff score is not necessarily a good indicator of disentanglement. Indeed, for 2D Shapes data, DIP-VAE-I has a higher Z-diff score  ($98.7$) and almost an order of magnitude lower reconstruction error than $\beta$-VAE for $\beta=60$, however comparing the latent traversals of $\beta$-VAE in Fig.~\ref{fig:sweeps_dsprites} and DIP-VAE-I in Fig.~\ref{fig:sweeps_dsprites_dipvaeI} indicate a better disentanglement for $\beta$-VAE for $\beta=60$ (though at the cost of much worse reconstruction where every generated sample looks like a hazy blob). On the other hand, we find the proposed SAP score to be correlated well with the qualitative disentanglement seen in the latent traversal plots. This is reflected in the higher SAP score of $\beta$-VAE for $\beta=60$ than DIP-VAE-I. 
We also observe that for 2D Shapes data, DIP-VAE-II gives a much better trade-off between disentanglement (measured by the SAP score) and reconstruction error  than both DIP-VAE-I and $\beta$-VAE, as shown quantitatively in Fig.~\ref{fig:recon-zdiff-dsprites} and qualitatively in the latent traversal plots in Fig.~\ref{fig:sweeps_dsprites}. The reason is that DIP-VAE-I enforces $[\text{Cov}_{p(x)} [\vmu_\phi(\vx)]]_{ii}$ to be close to $1$ and this may affect the disentanglement adversely by splitting a generative factor across multiple latents for 2D Shapes where the generative factors are much less than the latent dimension. For real datasets having lots of factors with complex generative processes, such as CelebA, DIP-VAE-I is expected to work well which can be seen in Fig.~\ref{fig:recon-zdiff-celeba} where DIP-AVE-I yields a much lower reconstruction error with a higher SAP score (as well as higher Z-diff scores).

{\bf Binary attribute classification for CelebA.~} We also experiment with predicting the binary attribute values for each test example in CelebA from the inferred $\vmu_\phi(\vx)$. For each attribute $k$, we compute the \emph{attribute vector} $\vw^k=\frac{1}{|\vx_i:y_i^k=1|}\sum_{\vx_i:y_i^k=1}\mu_\phi(\vx_i)-\frac{1}{|\vx_i:y_i^k=0|}\sum_{\vx_i:y_i^k=0}\mu_\phi(\vx_i)$ from the training set, and project the $\vmu_\phi(\vx)$ along these vectors. A bias is learned on these scalars (by minimizing hinge loss) which is then used for classifying the test examples. Table~\ref{tab:binclass} shows the results for the attribute which show the highest change across various methods (most other attribute accuracies do not change). The proposed DIP-VAE outperforms both VAE and $\beta$-VAE for most attributes. The performance of $\beta$-VAE gets worse as $\beta$ is increased further.

%% file: related.tex
\section{Related Work}
Adversarial autoencoder \citep{makhzani2015adversarial} also matches $q_\phi(z)$ (which is referred as \emph{aggregated posterior} in their work) to the prior $p(z)$. However, adversarial autoencoder does not have the goal of minimizing $\kl(q_\phi(\vz|\vx)||p_\theta(\vz|\vx))$ which is the primary goal of variational inference. It maximizes $\expect_\vx \left[\expect_{\vz\sim q_\phi(\vz|\vx)}\left[ \log p_\theta(\vx|\vz)\right]  \right]$ $- \lambda\,D(q_\phi(\vz)\Vert p(\vz))$, where $D$ is the distance induced by a \emph{discriminator} that tries to classify  $\vz\sim q_\phi(\vz)$ from $\vz\sim p(\vz)$ by optimizing a cross-entropy loss (which induces JS-divergence as $D$). This can be contrasted with the objective in~\eqref{eq:regelbo}. 

\noindent {\bf Invariance and Equivariance.~} Disentanglement is closely connected to invariance and equivariance of representations. If $R:\vx\to \vz$ is a function that maps the observations to the feature representions, equivariance (with respect to $T$) implies that a primitive transformation $T$ of the input results in a corresponding transformation $T'$ of the feature, i.e., $R(T(\vx))=T' (R(\vx))$. Disentanglement requires that $T'$ acts only on a small subset of dimensions of $R(\vx)$ (a sparse action). In this sense, equivariance is a more general notion encompassing disentanglement as a special case, however this special case carries additional benefits of interpretability, ease of transferrability, etc. Invariance is also a special case of equivariance which requires $T'$ to be identity for $R$ to be invariant to the action of $T$ on the input observations. However, invariance can obtained more easily from disentangled representations than from equivariant representations by simply marginalizing the appropriate subset of dimensions. There exists a lot of prior work in the literature on equivariant and invariant feature learning, mostly under the supervised setting which assumes the knowledge about the nature of input transformations (e.g., rotations, translations, scaling for images, etc.) \citep{schmidt12,Bruna,poggio2014inv,Anselmi_IAI,groupequiconvnets,dieleman-cyclic-2016,haarintkernels,mroueh2015,raj2016local}.

%% file: conclusion.tex
\section{Concluding Remarks}
We proposed a principled variational framework to infer disentangled latents from unlabeled observations. Unlike $\beta$-VAE, our variational objective does not have any conflict between the data log-likelihood and the disentanglement of the inferred latents, which is reflected in the  empirical results. We also proposed the SAP disentanglement metric that is much better correlated with the qualitative disentanglement seen in the latent traversals than the Z-diff score \cite{higgins2016beta}.   An interesting direction for future work is to take into account the sampling biases in the generative process, both natural (e.g., sampling the \emph{female} gender makes it unlikely to sample \emph{beard} for face images in CelebA) as well as artificial (e.g., a collection of face images that contain much more smiling faces for males than females misleading us to believe $p(\text{gender,smile})\neq p(\text{gender})p(\text{smile})$), which makes the problem challenging and also somewhat less well defined (at least in the case of natural biases).  Effective use of disentangled representations for transfer learning is another interesting direction for future work.

%% file: appendix.tex
\appendix
\begin{center}{\Large Appendix }\end{center}

% CelebA disentangle figures
\section{Latent traversals for 2D Shapes and Chairs dataset}
\begin{figure}[h]
\centering
\includegraphics[width=\textwidth]{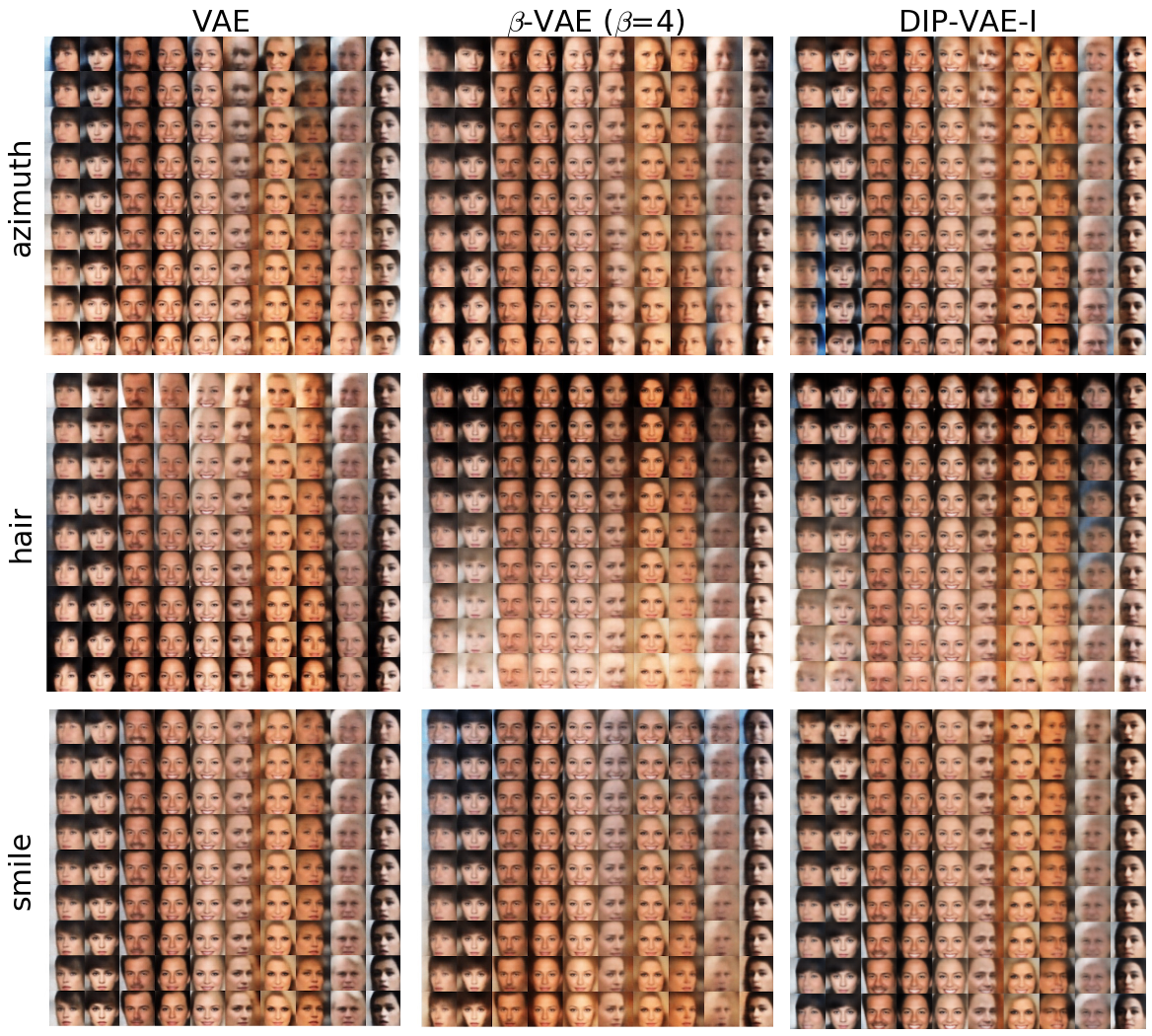}
\caption{Qualitative results for disentanglement in CelebA dataset.}
\label{fig:reconstruct_celeb}
\end{figure}

% Chairs disentangle figures
\begin{figure}[h]
\centering
\includegraphics[width=\textwidth]{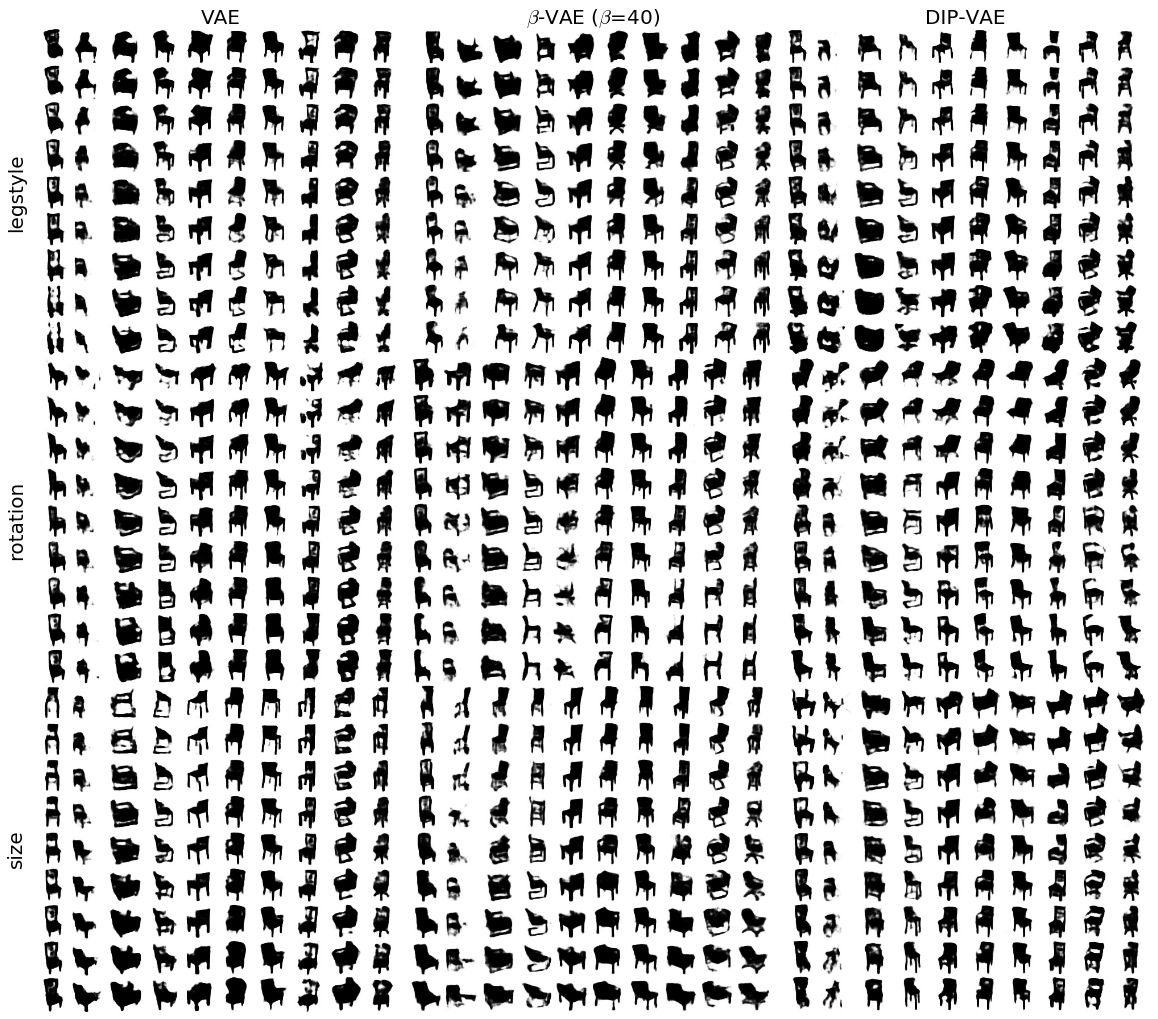}
\caption{Qualitative results for disentanglement in Chairs dataset.}
\label{fig:reconstruct_chairs}
\end{figure}